# An ensemble deep learning approach to detect tumors on Mohs micrographic surgery slides

Abdurrahim Yilmaz, MS[1,5]*, Serra Atilla Aydin, MD[2,3], Deniz Temur, BS[1], Furkan Yuceyalcin, BS[1], Berkin Deniz Kahya, BS[1], Rahmetullah Varol, MS[1], Ozay Gokoz, MD[2], Gulsum Gencoglan, MD[4], Huseyin Uvet, PhD[1,†], Gonca Elcin, MD[2,†]

[1] Yildiz Technical University, Mechatronics Engineering Department, 34349 Istanbul, Turkiye
[2] Hacettepe University, Department of Dermatology, 06230 Ankara, Turkiye
[3] Charité Universitaetsmedizin Berlin, Department of Dermatology, Venereology and Allergology, Berlin, Germany
[4] Istinye University, Department of Dermatology, 34408 Istanbul, Turkiye
[5] Imperial College London, Department of Metabolism, Digestion and Reproduction, Burlington Danes, Du Cane Rd, W12 0TN, London, United Kingdom
† Equally supervised

## ABSTRACT

Mohs micrographic surgery (MMS) is the gold standard technique for removing high risk nonmelanoma skin cancer however, intraoperative histopathological examination demands significant time, effort, and professionality. The objective of this study is to develop a deep learning model to detect basal cell carcinoma (BCC) and artifacts on Mohs slides. A total of 731 Mohs slides from 51 patients with BCCs were used in this study, with 91 containing tumor and 640 without tumor which was defined as non-tumor. The dataset was employed to train U-Net based models that segment tumor and non-tumor regions on the slides. The segmented patches were classified as tumor, or non-tumor to produce predictions for whole slide images (WSIs). For the segmentation phase, the deep learning model success was measured using a Dice score with 0.70 and 0.67 value, area under the curve (AUC) score with 0.98 and 0.96 for tumor and non-tumor, respectively. For the tumor classification, an AUC of 0.98 for patch-based detection, and AUC of 0.91 for slide-based detection was obtained on the test dataset. We present an AI system that can detect tumors and non-tumors in Mohs slides with high success. Deep learning can aid Mohs surgeons and dermatopathologists in making more accurate decisions.

# INTRODUCTION

Mohs micrographic surgery (MMS) is a precision-driven technique for treating non-melanoma skin cancers (NMSC). It employs a cyclic approach that begins with a minimal tumor excision followed by the mapping of the excised tissue and immediate intraoperative microscopic examination of the surgical margins. If remnant tumors are identified, additional excisions are carried out at the exact locations, as dictated by the map. This process repeats until all margins are clear of tumor cells, ensuring total eradication of the tumor while preserving as much healthy tissue as possible. The technique is not a histopathologic diagnostic tool for skin tumors, its value lies in its efficiency to track remaining tumor at the surgical margins, thus eliminating recurrences. Particularly effective against high-risk basal cell carcinoma (BCC), the most prevalent cancer globally, MMS has become the gold-standard treatment, further emphasizing its expanding necessity.

However, a critical factor that determines the success of MMS is the intraoperative histopathological examination, which is labor-intensive and time-consuming. This drawback can impede the broader adoption of MMS, potentially affecting patient outcomes. The rapid development of machine learning technology holds promise in streamlining the demanding tasks, including the microscopic examination of Mohs slides. Deep learning, a subfield of machine learning, has shown remarkable capabilities in histopathological image analysis, thereby offering prospects of automation and enhanced diagnostic and prognostic precision. Convolutional neural networks (CNNs), for instance, have demonstrated proficiency in tasks such as tumor detection, segmentation, and classification[1,2]. Moreover, recent strides in transfer learning have allowed for the successful adaptation of pretrained models to dermatology, resulting in better performance and expedited training times[3,4]. Generative models like generative adversarial networks (GANs) have also been deployed to mitigate the small labeled dataset in histopathology through data augmentation and image synthesis[5].

In spite of these advancements, existing literature on tumor and non-tumor area detection in MMS is sparse. Zon et al. pioneered a patch-based segmentation model using 171 slides from 70 patients, which achieved an area under the curve (AUC) value of 0.90 for slide-level classification of BCC[6]. Nevertheless, the study was limited by the sole reliance on a single expert for labeling with undetailed statistical analysis. Similarly, Sohn et al. conducted a study that yielded an AUC of 0.75, but it was hindered by the use of a relatively small dataset[7]. Thus, there is a pressing need for a more comprehensive dataset annotated by multiple experts to ensure the robustness of the resulting algorithm. Moreover, it is becoming increasingly clear that the development of robust and reliable AI systems can help alleviate the burden on experts and provide swift feedback for surgical procedures.

In light of these considerations, the present study aims to develop an AI model capable of accurately detecting tumor and non-tumor regions during the histopathological examination of MMS using an ensemble learning approach. We leveraged U-Net based deep learning models, known for their precision, to detect tumor and non-tumor regions in MMS images. We trained these models using a dataset of 731 high-resolution MMS images including 91 tumor and 640 non-tumor annotations. Additionally, we computed specific metrics to assess the model's performance and reliability. The overarching intent of this study is to cultivate a synergy between human expertise and AI in histopathology to optimize the benefits and efficiency of MMS.

## RESULTS & DISCUSSION

In this study, we present an ensemble learning system based on deep neural networks for tumor and artifact detection on histopathological slides of MMS. We evaluated the performance of our U-Net-based deep learning models. Our approach demonstrated remarkable success, as quantified by the Dice score and AUC. The Dice score, a metric of overlap between the predicted and ground truth segmentation, was used to assess the performance of our model. In the test set evaluations, our model achieved a Dice score of 0.70 and 0.67 for tumor and non-tumor segmentation, respectively, alongside AUC scores of 0.98 and 0.96. These results indicate a high degree of overlap between the model's predictions and the ground truth. Comparatively, a state-of-art autoML segmentation method, nnU-Net, has a 0.64 Dice and 0.83 AUC score. These results underline the effectiveness of our model in accurately detecting tumor and non-tumor regions within histopathological images. For patch and slide classification, our model achieved 0.98 and 0.91 AUC scores on the test set for tumors, respectively. These scores indicate a high level of success in distinguishing tumor and artifact regions from healthy tissue and each other. The results of the study are shown in Table 2 and Figure 2 with example images.

Our study demonstrates the potential of using a deep learning system to assist clinicians for histopathological analysis of MMS slides. MMS is widely recognized as an extremely effective treatment for high risk BCCs and SCCs, with the highest reported cure rates[12]. It exhibits a remarkable 5-year cure rate of up to 99% for first-time treated cancers and 94% for recurrent cancers[13]. The success of MMS creates an upward trend in the use of MMS for the treatment of over 1 million annual skin cancer diagnoses in the United States alone. The main reason for this trend is that BCC incidence rates are rising globally due to factors such as an aging population and increased exposure to UV radiation, leading to a staggering lifetime risk of 30% in the US[14,15]. With the rising global incidence of BCC, the workload of clinicians will be increasing exponentially.

This growing burden on clinicians underscores the need for more widespread expertise in Mohs surgery. However, its training is conducted in a very limited number of centers and requires a long training period. Furthermore, the most significant challenge in this field is learning to accurately assess the tumor in pathology specimens rather than the surgical removal. In this context, the application of deep learning systems can be particularly transformative.

Addressing these challenges, AI can ensure that the evaluation process during the surgical operation is conducted accurately and effectively. The results of our study suggest that Mohs surgeons may benefit from AI in challenging cases where there is uncertainty about the presence of a tumor. Incorrectly assessing non-tumor regions as tumors may lead to the unnecessary removal of healthy tissue, causing functional and cosmetic losses.

To improve the model's accuracy in detecting tumor regions, we trained the AI using high-magnification patches from digitized WSI of Mohs slides. This approach aligns with clinical workflows, where pathologists begin with a low magnification overview and then shift to high magnification for detailed examination. By focusing on high-magnification patches, the AI can better identify subtle morphological changes critical for distinguishing tumor cells from surrounding tissue. Additionally, a tissue detection algorithm was employed to automatically identify and isolate tissue-containing regions, enabling comprehensive slide analysis while minimizing time spent on non-informative areas. This combination of targeted training and efficient tissue detection enables the automated analysis of a Mohs WSI.

Our study supports the idea that training the AI model should be conducted not only for the tumor but also for potential non-tumor regions that can be confused with the tumoral area. The segmentation of non-tumor regions, including artifacts, can play a crucial role in enhancing the performance of AI models for histopathological image analysis. An ideal Mohs histopathologic section should be devoid of any artifacts such as fragmentation, folding, bubbles or holes. However, real-life experience rarely allows perfectly ideal slides. To address this, we trained our model also on slides with pronounced artifacts that, while not meeting the quality assessment standards, were valuable for training the model to recognize artifacts. The inclusion of such slides and artifact segmentation as a part of the ensemble approach aimed to strengthen the model's ability to distinguish between tumors and artifacts, thereby preventing false positive cases where artifacts could be mistaken for tumors. In our study, we chose a full image training approach to detect non-tumor regions. This decision was based on the understanding that some non-tumor artifacts have more complex shapes, and preserving their spatial information is essential, which patching might compromise. The automatic detection of non-tumor artifacts can also be part of a quality control system, ensuring the integrity of histopathological images. Such systems can potentially save examiners valuable time by alerting them to potential issues with a slide before they commence their analysis[16]. Furthermore, certain non-tumor artifacts might be mistaken for pathological features, leading to potential misinterpretations and incorrect diagnoses. Given that many machine learning algorithms are sensitive to these artifacts, their mismanagement can negatively affect the success of machine learning models if they are not correctly handled. In addition to non-tumor artifact detection, the annotation of artifacts is also more challenging because artifacts such as water droplets can be placed both on tumor and healthy tissue. This situation elevates the risk of false positive and negative rates in AI model training. Accurate annotation, detection, and segmentation of these artifacts are crucial to mitigating these risks. Moreover, this process can be used as examples during the training of residents and fellows to demonstrate common pitfalls and issues in slide preparation. Thus, while the primary focus in digital pathology often lies in the segmentation of biologically relevant features such as tumors, the identification of non-tumor artifacts is equally vital to develop successful AI systems for clinical practice.

There are significant opportunities for the application of AI in histopathology that offers huge potential. The automation of tasks traditionally performed by pathologists can improve diagnostic success, reduce workload, and provide faster feedback for surgeries[17]. This could, in turn, lead to a reduction in healthcare costs and improved patient outcomes[18]. Building on these advancements, we have developed our model with superior performance compared to previous research to detect tumors in MSS. Previous studies conducted by Zon et al. and Sohn et al. made important contributions to this field[6,7]. Recently, a study by Tan et. al proposed a tile-based classification model to produce probability maps as overlay to the Mohs focused microscopic views and validated their model in a prospective setting[19]. Our study distinguishes itself from prior works in several ways. We utilized two U-Net based models for tumor and non-tumor artifact segmentation followed by a classification model. Our study benefits from an extensive and diverse dataset, which led to adequate model generalization in our internal test set.

Nevertheless, our study is not without limitations. The annotations were made by two experts and could be subject to individual variations and biases. Future studies could aim to incorporate a consensus approach to annotation, where multiple experts independently annotate the images, and a consensus is reached on the final annotations. In addition, although our model showed promising results, its clinical applicability needs to be validated through prospective studies in a real-world clinical setting. Furthermore, while our model focused on BCC, future works could expand its training set to include other NMSC cases such as squamous cell carcinoma (SCC) to extend the models applicability to different tumor types. Additionally, due to the rarity of mimicker lesions and incidental malignancies in our dataset, our AI model was not specifically trained to differentiate BCC from benign entities commonly encountered during histopathological examination of Mohs slides (e.g., basaloid follicular hamartoma, granulomatous inflammation, syringoma) or incidental malignant lesions (e.g., atypical fibroxanthoma, sebaceous carcinoma, Merkel cell carcinoma)[20]. While this constitutes a limitation of our dataset and hence the resulting AI model, future studies could incorporate sufficient examples of these entities into their datasets to discriminate between BCC and its histopathologic mimickers.

In conclusion, our study demonstrates the potential of a U-Net based model in histopathological image analysis for both tumor and non-tumor segmentation. This promising technology could have important impacts on pathology, transforming it from a predominantly manual practice to an automated one, enhanced by the precision and efficiency of AI. While considerable progress has been made, the field remains ripe for further exploration. Future research could investigate the identification and classification of different types of skin cancers with telepathology. Deep learning and decision support systems could be developed to increase the success of tumor detection and accelerate workflow for slide analyses.

## METHODS

**Dataset**

The dataset utilized in this retrospective study is composed of hematoxylin and eosin (H&E) stained histopathological slides obtained from the MMS procedures performed in the MMS Unit of Hacettepe University Faculty of Medicine Department of Dermatology by an ESMS certified Mohs Surgeon (G.E.) between 2013 and 2020. Mohs slides were stored in the archives of the Department of Pathology, all of which underwent routine intraoperative quality assessment by the reporting dermatopathologist (O.G.) to ensure high quality of slides for complete surgical margin evaluation. For AI model development, slides of varying quality, including those with pronounced artifacts, were included to increase robustness. These H&E stained frozen section histopathology slides that were obtained from MMS were digitized with Olympus VS120 digital slide scanner at a 40x magnification, anonymized, and saved in .vsi format (S.A.). As a result, a total of 731 high resolution whole slide images were acquired. Of these 731 images, images that belong to 51 patients were included in this study. AI engineer authors (A.Y., D.T., F.Y., B.D.K., R.V.) manually reviewed all slides in the dataset to ensure they meet the necessary standards such as image clarity for AI model training. The initial dataset consisted of very high-resolution images with a pixel size of 0.3939 $\mu$m. Average width and height of slides were 83621 and 46123 pixels, respectively. Average and total file size were 11.92 GB and 2,085 GB, respectively. Ethical approval was obtained from the Institutional review board of Hacettepe University (reference number: GO 21/1178, approval date: 05 April 2022).

Annotation of the images was performed by a senior dermatologist, certified in MMS (G.E.) and a senior dermatopathologist (O.G.) who identified and labeled the tumor (red color) and non-tumor regions (green color) using QuPath (version 0.4.2)[20], a digital pathology program. The diverse expertise of the annotating pathologists contributes to the comprehensiveness and accuracy of the annotated dataset.

For training of the deep learning model, 731 high-resolution annotated slides were cropped into smaller sizes to omit empty regions and increase model performance, yielding 91 tumor regions and 640 non-tumor regions. The dataset specifications are shown in Table 1. Resolutions of each Mohs surgery slide were downgraded with a scale of 2X to obtain best performance[8]. Experts (G.E. and O.G.) annotated tumor regions in red color and non-tumor regions in green color across the entire Mohs sections. After annotation, these areas were extracted and saved as mask images in separate .png files. The green channel was used to filter non-tumor regions, and the red channel was used to filter tumor regions. These filtered masks were then converted to grayscale and subsequently to binary images to reduce computational load. Then, the non-tumor images were resized to 1024 x 2048 pixels. Finally, pixel values of mask images and the corresponding histopathology images were normalized between 0 and 1 values. These regions, along with their labels, were used as ground truth during training and validation of our deep learning model. To enhance the generalizability of the model and prevent overfitting, data augmentation techniques were applied to the patches using random rotations, horizontal and vertical flips, and zooms. Overview of the study is shown in Figure 1.

**Deep Neural Networks and Model Training**

For an efficient segmentation of tumor and artifacts, we present an ensemble approach based on ensemble learning and U-Net[9]. To detect tumors and non-tumor regions, we deployed two different U-Net architectures, a type of CNN designed specifically for image segmentation tasks. The architecture consists of an encoding path that successively reduces the spatial dimensions of the input image while increasing the number of feature channels, and a decoding path that restores the spatial dimensions while combining the high-level features with the corresponding low-level features from the encoding path with their corresponding position on images. The final layer of the U-Net model outputs a segmented image, where each pixel is assigned a class label (tumor or non-tumor region).

For model training, we used the capabilities of TensorFlow library (version 2.10) to facilitate a streamlined data processing workflow on Python programming language (version 3.9.5), which allowed us to efficiently manage the dataset. To prepare for model training and evaluation, we divided the labeled data into three sets: the training, validation, and test sets. The training set comprises 70% of the labeled data, while the validation set and test set each constitute 15% of the dataset. For segmentation models, ResNet-34 architecture which is pre-trained on ImageNet dataset was used as a down-sampling part of U-Net architecture[10]. The segmentation models were optimized using Adam optimization algorithm with initial learning rate of 2e-4 and ReduceLRonPlateu approach that changes learning rate with its value of 0.2 when metrics do not increase for five epochs. The performance of the U-Net model was evaluated on a separate test dataset of image patches. For training and optimization of DNN, a workstation with Nvidia GeForce RTX3090, Intel i7-11700KF, and 128 GB Ram was used.

The first model is for the non-tumor segmentation. To preserve underlying structures and ensure efficient processing, we resized all images to a consistent resolution of 1024 x 2048 pixels. This resizing approach was chosen to find a balance between maintaining essential anatomical details and avoiding computational bottlenecks.

The second model is for the tumor segmentation model training, the patches of 256 x 256 pixels were extracted from original images and corresponding labels. The original image dimensions remained unchanged to preserve fine-grained details. This patch-based strategy not only preserved crucial details but also enabled effective handling of complex and varying tumor patterns while helping with the computation cost that often accompanies the analysis of large histopathology images. As part of our training strategy, we implemented a tissue detection algorithm, where 80% of the non-tissue patches were randomly excluded.

The third model is for training the tumor classification model, the same patches of 256 x 256 pixels as for the tumor segmentation model were used. This labeled dataset was trained on the ResNet-101 architecture, a neural network that had been pretrained on the ImageNet dataset.

In addition to our custom non-tumor segmentation methods, we leveraged the nnU-Net frameworks capabilities. nnU-Net, a self-configuring method for deep learning-based biomedical image segmentation, proved to be a valuable component in our research pipeline. The nnU-Net framework, designed by Isensee et al.[11], operates as a versatile tool for automating the development of segmentation models in the context of medical imaging. By employing U-Net architecture as its base, nnU-Net dynamically adapts its architecture and hyperparameters based on the characteristics of the input data. This unique feature eliminates the need for laborious manual tuning, streamlining the process of model development.

**Statistical Analysis**

To measure segmentation success, Dice similarity coefficient (Dice score) and AUC score were calculated. The Dice coefficient is a quantitative statistic utilized to assess the similarity between two sets. The Dice score is a metric used to quantify the degree of overlap between two sets of objects. We also evaluated the performance of our model using the AUC score, which measures the model's ability to distinguish between positive and negative regions. An AUC score of 1 denotes perfect discrimination, while a score of 0.5 indicates a performance no better than random chance.

## Acknowledgments

Abdurrahim Yilmaz has been funded by the President's PhD Scholarships at Imperial College London.

## Data Availability Statement

The datasets used and/or analyzed during the current study are available from the corresponding author on reasonable request.

## Additional Information

There is no competing interest.

## Author contributions statement

Conceptualization, A.Y., S.A.A, G.G., H.U., and G.E.; Methodology, A.Y., S.A.A., D.T., F.Y., B.D.K. and R.V.; Data Collection, S.A.A. and O.G.; Writing–Original Draft Preparation, A.Y., S.A.A., H.U. and G.E.; and A.Y. and R.V. prepared figures. All authors reviewed the manuscript.

# Tables

Table 1: Dataset characteristics

| Attribute | Amount |
|---|---|
| Total Patients | 51 |
| Total Crops | 731 (100%) |
| Crops with Artifacts | 640 (88%) |
| Crops with Tumours | 91 (12%) |

Table 2: Performance metrics for the segmentation and classification models in tumor and artifact analysis.

| Type | Model | AUC Score | Dice Score |
|---|---|---|---|
| Tumor Segmentation | U-Net (ResNet-34) | 0.98 | 0.70 |
| Artifact Segmentation | nnU-Net | 0.83 | 0.64 |
| Artifact Segmentation | U-Net (ResNet-34) | 0.96 | 0.67 |
| Patch-Level Tumor Classification | ResNet-101 | 0.98 | - |
| Slide-Level Tumor Classification | ResNet-101 | 0.91 | - |

# Figures

Fig. 1. The workflow for tumor segmentation on histopathology slides starts with collecting skin tissue from a patient, scanning it to obtain whole slide images (WSIs), and annotating the images by experts. The annotated images were preprocessed and the patches were extracted. A U-Net artifact detection model segments artifacts in the input images, and a U-Net tumor detection model further processes the patches to segment tumor regions. The final output is a segmented image highlighting tumor areas (red) and artifacts (green).

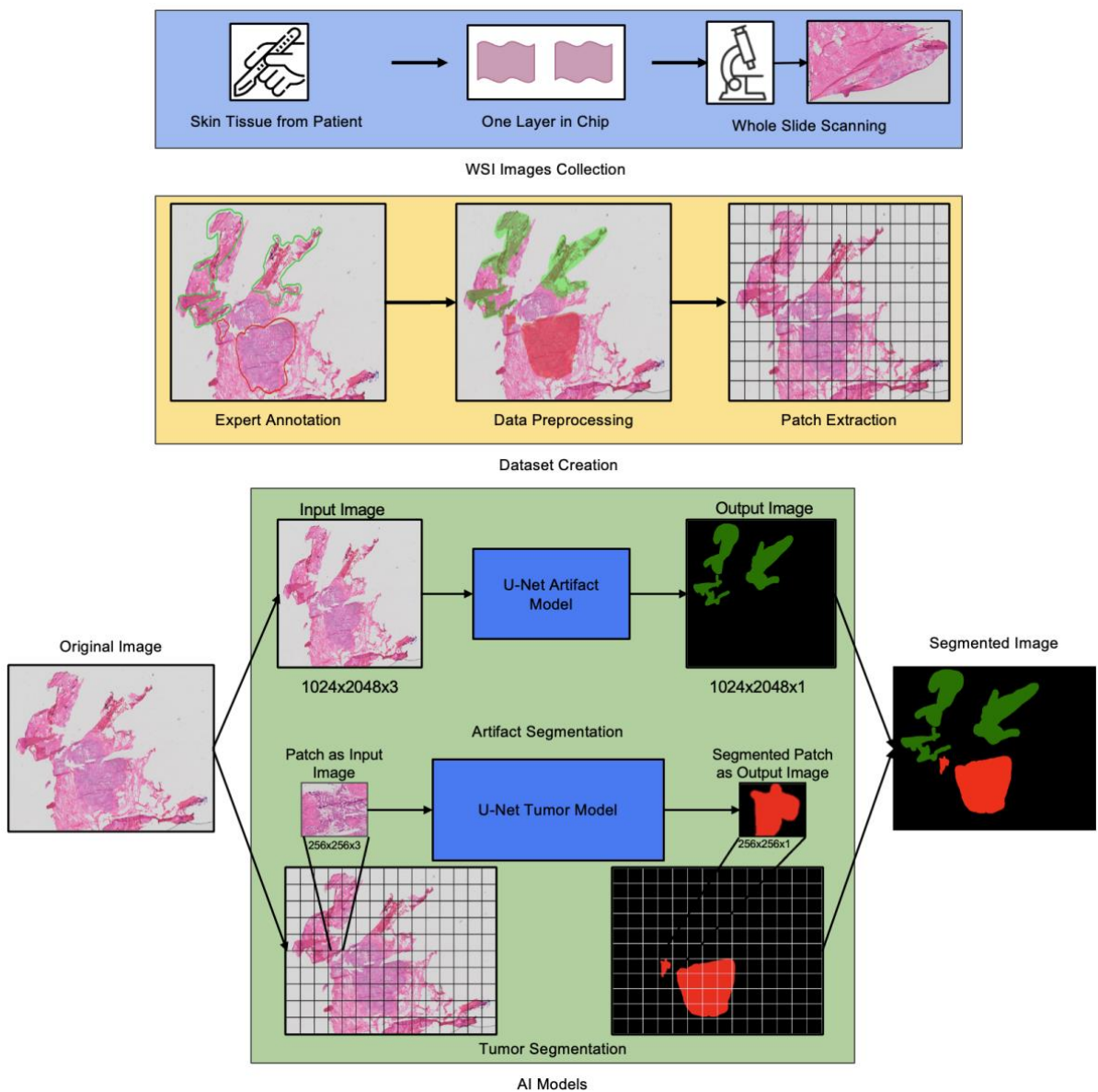

Fig. 2. Comparison of ground truth masks and predicted masks for tumor and artifact segmentation on Mohs slides. The top and middle rows display original images, corresponding ground truth masks, and predicted masks. The bottom panel shows the ROC curve for tumor classification, indicating a high area under the curve (AUC) of 0.98.

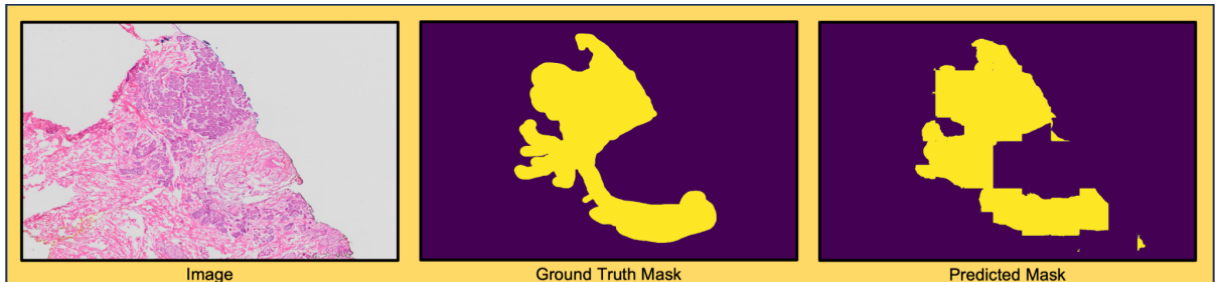

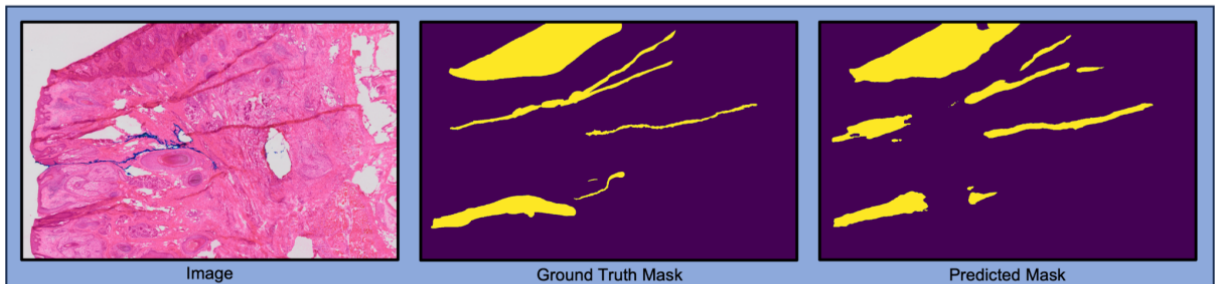

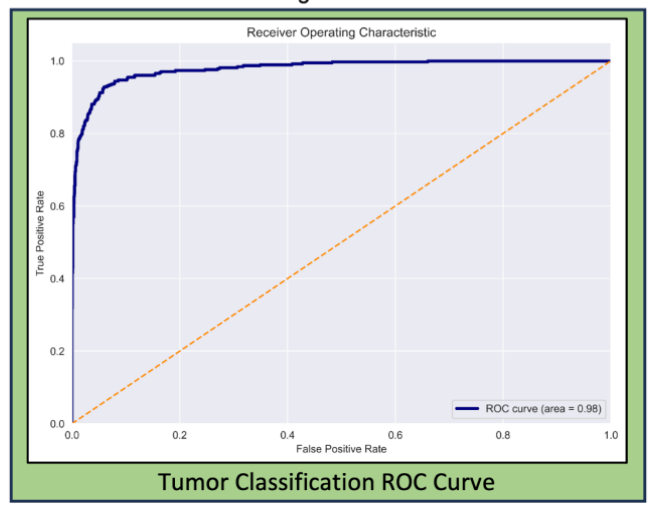